\documentclass[11pt]{article}
\usepackage{acl}
\usepackage{comment}
\usepackage{graphicx} 
\usepackage{multirow}

\usepackage{times}
\usepackage{latexsym}
\usepackage[T1]{fontenc}
\usepackage[utf8]{inputenc}
\usepackage{microtype}
\usepackage{inconsolata}

\usepackage{amssymb}
\usepackage{amsmath}
\usepackage{mathtools}
\usepackage{amsthm}
\usepackage{float}
\usepackage{longtable}
\usepackage{paralist}

\definecolor{brickred}{rgb}{0.8, 0.25, 0.33}

\title{Little Giants: Exploring the Potential of Small LLMs as Evaluation Metrics in Summarization in the Eval4NLP 2023 Shared Task}
\author{Neema Kotonya  \quad Saran Krishnasamy \\
        \textbf{Joel Tetreault} \quad  \textbf{Alejandro Jaimes}\\ 
        Dataminr Inc. \\
\texttt{\{neema.kotonya,skrishnasamy,jtetreault,ajaimes\}@dataminr.com}}
\date{September 2023}

\begin{document}

\maketitle

\begin{abstract}

This paper describes and analyzes our participation in the 2023 Eval4NLP shared task, which focuses on assessing the effectiveness of prompt-based techniques to empower Large Language Models to handle the task of quality estimation, particularly in the context of evaluating machine translations and summaries. We conducted systematic experiments with various prompting techniques, including standard prompting, prompts informed by annotator instructions, and innovative chain-of-thought prompting. In addition, we integrated these approaches with zero-shot and one-shot learning methods to maximize the efficacy of our evaluation procedures. Our work reveals that combining these approaches using a ``small'', open source model (\verb|orca_mini_v3_7B|) yields competitive results.

\end{abstract}




\section{Introduction}
Large Language Models (LLMs) have revolutionized the field of Natural Language Processing (NLP) by demonstrating remarkable proficiency in a multitude of generative tasks~\cite{llmfewshot}. Beyond their capabilities in text generation, LLMs offer the potential to automate the evaluation of generated text, particularly in domains such as machine translation and summarization.

Previous research efforts have explored LLM-based evaluation metrics, yielding promising results. Notable examples include the development of metrics like the GEMBA metric for translation quality assessment~\cite{kocmi-federmann-2023-large}, work on the effectiveness of LLMs as an alternative to human evaluation for NLP tasks by \citet{chiang-lee-2023-large}, and the INSTRUCTSCORE metric for summarization evaluation~\cite{xu2023instructscore}. However, a significant gap exists in the systematic evaluation and exploration of  
prompting techniques available for metric usage with LLMs.  In fact, there is scant work in this area to date. Exceptions include the work of \citet{mendonca-etal-2023-simple} for dialogue evaluation, \citet{yang2023knowledgeprompted} and GEMBA for MT evaluation, \textsc{G-Eval}, a chain-of-thought based framework for the evaluation of generated texts that leverages GPT-4 \citep{DBLP:journals/corr/abs-2303-16634}, and \textsc{GPTScore} for text generation evaluation \cite{DBLP:journals/corr/abs-2302-04166}.  



This paper presents our contribution to addressing this gap through our participation in the \textit{Prompting Large Language Models as Explainable Metrics} shared task~\cite{eval4nlp23}, which was conducted as part of The 4th Workshop on Evaluation \& Comparison of NLP Systems, hosted at AACL 2023. We delve into various prompting approaches and techniques, offer a comprehensive overview of the results we have obtained, and provide a thorough analysis of our findings (our team name is \textsc{CompetitionEntrants}).



We exclusively participated in the Small Models Track, focusing on models with parameters less than or equal to 25 billion, for the Summarization task. Consequently, all our experiments and reported results are derived from the \verb|orca_mini_v3_7B| model. Among our various approaches, the best-performing one, employing a standard prompt in a zero-shot setting, achieved a score of 0.44 during the test phase of the shared task. While this performance is notable, it places us marginally behind the leaderboard's highest score of 0.50 by a margin of 0.06.  We also find that Chain-of-Thought (CoT) prompting~\cite{chainofthought} also aids in explicating the evaluation task to the model. This technique holds the potential to enhance the interpretability and explainability of quality estimation models.


\section{Task Description}

The primary objective of the shared task is to investigate prompt-based methodologies for LLMs in the development of automated quality metrics in a reference-free setup tailored to natural language generation tasks, specifically summarization and machine translation. 
These quality scores are constructed using fine-grained scoring or error labels. The underlying rationale behind reference-free evaluation metrics is to provide assessment scores that are comparable to those of reference-based metrics while reducing reliance on often noisy and costly reference labels.  A typical reference-free metric operates by taking a source (SRC) and a hypothesis (HYP) as inputs and subsequently generates a metric score, thereby providing an alternative to traditional reference-based evaluation methodologies.

\subsection{Task Setup}

The shared task is structured into two distinct sub-tasks, each contributing to the overarching goal:

\begin{enumerate}
\item Prompting Strategies for Evaluation of Summarization and Translation:
This entails designing prompts and instructions that enable the assessment of the quality of generated content. The outcome of this sub-task serves as a critical component in the reference-free evaluation process.

\item Score Aggregation:
The second sub-task focuses on the creation of a score aggregator mechanism. The primary objective here is to devise a method that computes an overall quality estimation score based on the outputs generated in Task 1. This aggregator consolidates individual quality assessments into a comprehensive quality estimation, ultimately providing a holistic evaluation of translation and summarization performance.
\end{enumerate}

\subsection{Datasets}

Data is provided for the tasks of summarization and machine translation:

\begin{itemize}
    \item Summarization: The training and development data for this track is derived from the datasets detailed in SummEval~\cite{fabbri2020summeval} with the scores being the average of human annotations across the four aspects - coherence, consistency, fluency, and relevance. 
    \item Machine Translation: The training and development datasets are derived from the Multidimensional Quality Metrics (MQM) annotations of the WMT22 metrics shared task \cite{freitag-etal-2022-results} for machine translation.
\end{itemize}
For the test data, we are provided with a new reference-free dataset with sentence/summary-level quality scores for summarization and MT. As part of the test phase, 2 new language pair datasets, English-Chinese and English-Spanish are introduced for the machine translation track.
Table \ref{table:dataset} shows the counts of the train, development, and test datasets.

\begin{table}[hbt!]
\begin{tabular}{|cl|lll|}
\hline
\multicolumn{2}{|c|}{\textbf{task}  }           & \multicolumn{1}{|c}{\textbf{train}} & \textbf{dev} & \multicolumn{1}{c|}{\textbf{test}} \\ \hline
\multicolumn{2}{|l|}{Summarization}    & 320  &  1,280   &  825     \\ \hline
\multirow{4}{*}{Translation} & en-de & 11,046  & 7,364    &    1,425  \\
                             & zh-en & 15,750  & 10,500   &     -    \\
                             & en-es & -      &  -      &    1,834  \\
                             & en-zh & -      &  -      &    1,297  \\\hline          
\end{tabular}
\caption{ Train, Dev, and Test dataset sizes for summarization and machine translation tasks. Entries with - indicate that the dataset wasn't provided as part of this task.}
 \label{table:dataset}
\end{table}

\subsection{Large language models}
The following six Huggingface LLMs were permitted for use in the shared task, 
two larger models (65B and 70B parameter models), which we denote with the following symbol $\bullet$, and four smaller models, denoted by $\star$, each of which has 13B parameters or fewer: 

\begin{compactenum}

\item \texttt{Guanaco-65B-GPTQ}$^\bullet$\footnote{\href{https://huggingface.co/TheBloke/guanaco-65B-GPTQ} {https://huggingface.co/TheBloke/guanaco-65B-GPTQ}}
    \item \texttt{WizardLM-13B-V1.1-GPTQ}$^\star$\footnote{\href{https://huggingface.co/TheBloke/WizardLM-13B-V1.1-GPTQ}{https://huggingface.co/TheBloke/WizardLM-13B-V1.1-GPTQ}}
    \item \texttt{Nous-Hermes-13b}$^\star$\footnote{\href{https://huggingface.co/NousResearch/Nous-Hermes-13b}{https://huggingface.co/NousResearch/Nous-Hermes-13b}}
    \item \texttt{Platypus2-70B-Instruct-GPTQ}$^\bullet$\footnote{\href{https://huggingface.co/TheBloke/Platypus2-70B-Instruct-GPTQ}{https://huggingface.co/TheBloke/Platypus2-70B-Instruct-GPTQ}} 
    \item  \texttt{OpenOrca-Platypus2-13B}$^\star$\footnote{\href{https://huggingface.co/Open-Orca/OpenOrca-Platypus2-13B}{https://huggingface.co/Open-Orca/OpenOrca-Platypus2-13B}} 
    \item  \texttt{Orca\_mini\_v3\_7b}$^\star$\footnote{\href{https://huggingface.co/pankajmathur/orca_mini_v3_7b}{https://huggingface.co/pankajmathur/orca\_mini\_v3\_7b}}
 
\end{compactenum}



Despite having access to these LLMs, our work faced computational constraints that influenced our choice of models for experimentation. As a result, we focused primarily on experimenting with two small LLMs: \verb|orca_mini_v3_7B model| and \verb|Nous-Hermes-13b|. During the submission phase to the shared task's leaderboard, the final test results we presented were exclusively derived from the \verb|orca_mini_v3_7B| model. The shared task guidelines explicitly forbade model fine-tuning. 
\subsection{Evaluation}
For the evaluation process, we used Codalab as the platform for submitting our system entries. Notably, the organizers of the evaluation, as detailed by \citet{kocmi-federmann-2023-large}, provide direct assessment baselines for these LLMs. These baselines serve as reference points for evaluating the performance of our system and other participants in the shared task.

To quantify the performance of our system and the competing teams, the shared task organizers utilized the Kendall rank coefficient as the evaluation metric. The Kendall rank coefficient stands as an alternative to more traditional correlation metrics like Pearson's r and Spearman's $\rho$ correlations. It finds particular utility in situations where the data fails to meet specific assumptions or when dealing with relatively small sample sizes.

\section{Approaches} \label{sec:Approaches}


Three main classes of strategies are employed to enhance prompt effectiveness and interpretability for evaluating generated summaries. The first strategy, Core Prompts, encompasses three one-step methods for generating prompts.  The first two borrow from existing literature, and the final uses an LLM to simply generate a prompt from scratch.  Next, in Section \ref{ssec:promptrefine}, we introduce three methods (one manual, two automatic) to take prompts in Section \ref{ssec:coreprompts} and further refine them.  This is akin to paraphrasing in bulk.  Finally, in Section \ref{ssec:inference}, we outline two simple approaches for further refining prompts generated earlier. In total, the three classes span 8 different approaches, and approaches can be combined across the classes.

\subsection{Core Prompts} \label{ssec:coreprompts}

\paragraph{(1) Standard Prompting:}
Our initial approach was to formulate prompts (a total of 9 prompts) 
that assess summary quality across the four dimensions outlined in \citet{kryscinski-etal-2019-neural}: fluency, coherence, consistency, and relevance. These prompts task the model with generating quality scores for summaries, 
for different score ranges. In Table \ref{tab:standard-prompting}, we provide an example of standard prompting. In this example, the prompt specifies that the summarization should be rated from 1 and 5, with increments of 0.5 permitted.


\begin{table}[H]
\begin{tabular}{|p{7.2cm}|}
\hline
Given the following summary for a news article, evaluate this summary for its fluency, coherence, consistency and relevance. Provide an overall score for the quality of this summary in the range 1 (worst) to 5 (best). Possible scores are 1, 1.5, 2, 2.5, 3, 3.5, 4, 4.5 and 5.\\
\hline
\end{tabular}
\caption{
An example of a standard prompt for summarization quality evaluation that stipulates scores should be in the range 1 to 5, and intermediate scores should be in 0.5 intervals.}
\label{tab:standard-prompting}
\end{table}

\noindent \textbf{(2) Annotator Instructions as Seed Prompts:} To facilitate summarization evaluation, we then employed the instructions provided to expert annotators in \citet{fabbri2020summeval} (See Table \ref{tab:seed-prompting}).  These served as foundational ``seed prompts'' for subsequent prompt refinement. We conducted an assessment of this seed prompt by utilizing a subset of examples from the Eval4NLP training dataset, noting that these instructions exhibit relatively favorable performance on the training data.  

\begin{table}[H]
\begin{tabular}{|p{7.2cm}|}
\hline
[$\hdots$] In this task you will evaluate the quality of summaries written for a news article \\
To correctly solve this task, follow these steps: \\
1. Carefully read the news articles, be aware of the information it contains. \\
2. Read the proposed summary. \\
3. Rate each summary on a scale from 1 (Worst) to 5 (Best) by its relevance, consistency, fluency, and coherence. \\
Relevance: ``The rating measures how well the summary captures the key points of the article. \\
Consider whether all and only the important aspects are contained in the summary.'' [$\hdots$] \\
\hline
\end{tabular}
\caption{
Instructions provided to expert annotators in ~\citet{fabbri2020summeval}.}
\label{tab:seed-prompting}
\end{table}

% ...

\paragraph{(3) Prompt Generation via LLMs:} Additionally, we employed a separate LLM to generate a prompt entirely from scratch. The intuition behind this approach is that an LLM-derived prompt may yield improved results over a manually-crafted prompt. Similar intuition is followed in previous works that use LLMs to produce high-quality labels for LLM-generated texts \cite{zellers2019defending,DBLP:journals/corr/abs-2302-04166}.
For prompt generation we author simple prompts, instructing the LLM to generate a sequence of instructions based on the requirements (score range and aspects to consider for summarization) specified in the prompt. (see Table \ref{tab:prompt-instructions}).


\begin{table}[H]
\begin{tabular}{|p{7.2cm}|}
\hline
Write a set of instructions to evaluate the quality of the summary of a news article according to its coherence, consistency, fluency, and relevance for each sentence in the summary with respect to the news article. Each aspect (coherence, consistency, fluency, and relevance) should be scores from 1 to 5. 1 is the worst possible score, 5 is the best possible score. Instructions:\\
\hline
\end{tabular}
\caption{Instructions for prompting LLMs to generate prompts for summarization quality evaluation.}
\label{tab:prompt-instructions}
\end{table}



\subsection{Prompt Refinement} \label{ssec:promptrefine}

To further enhance the prompts' quality and effectiveness from Section \ref{ssec:coreprompts}, we employ three key strategies, one manual and two automatic:

\paragraph{(4) Manual Prompt Rewriting:}
This method involves meticulous manual rewriting of the instructions (done by the authors). We created prompt variations to elicit fine-grained answers, seek explanations for the provided answers, and employ templates specifying the desired answer format. These steps ensure that the instructions are refined to enhance the clarity and comprehensibility of the prompts. We also experiment with prompts that instruct the LLM to output both scores and explanations, similar to other works that explore both prediction (which in our case is quality estimation) and explanation jointly \cite{camburu2018snli,kotonya-toni-2020-explainable,chainofthought}. However, we find that prompting for explanations in addition to quality estimation yields poor results (see Table \ref{tab:prompt-explanation} in Appendix \ref{appendix:explanations}).

\paragraph{(5) Instruction Enhancement via LLMs:}
In this strategy, we provide the seed prompt as context and prompt a separate LLM to enhance the existing instructions. We utilize various phrases such as ``Improve the following instructions'', ``Rewrite the following instructions to yield better responses'', ``Write a more precise set of instructions'', and ``Rewrite the instructions below in order to yield the best results '' (see Table \ref{tab:prompt-refinement-example}).

\begin{table}[H]
\begin{tabular}{|p{7.2cm}|}
\hline
\verb|### System:| You are an AI assistant that follows instruction extremely well. Help as much as you can. \newline
 \verb|### User:|\newline
{\color{blue}Improve the following instructions:} \newline
``In this task you will evaluate the quality of summaries written for a news article\newline
To correctly solve this task, follow these steps:\newline
1. Carefully read the news article, be aware of the information it contains.\newline
2. Read the proposed summary.\newline
3. Rate each summary on a scale from 1 (Worst) to 5 (Best) by its relevance, consistency, fluency, and coherence.$\hdots$''\newline
{\color{blue} New instructions:}\\
\hline
\end{tabular}
\caption{Example of a prompt supplied for LLM-based prompt refinement, where the instruction used is ``Improve the following instructions.''}
\label{tab:prompt-refinement-example}
\end{table}




\paragraph{(6) Chain-of-Thought (CoT) Prompting:} We also harness the CoT prompting technique, which guides the model's evaluation process through a sequence of intermediate reasoning steps leading to the determination of the quality score for the provided summary. The main advantages of CoT prompts are that their specificity should reduce the number of inconsistencies in the generated response, yield responses that correlate strongly with human judgments and also allow for more complex reasoning.  The CoT additions are marked in blue in Tables \ref{tab-cot1} and \ref{tab-cot2}.
\newline \\



\subsection{Exploration of Inference Settings} \label{ssec:inference}
Given the constraint of not permitting model fine-tuning, we explore various inference settings to optimize model performance:

\paragraph{(7) Zero-shot Approach:} In this setting, the model is evaluated and prompted to generate responses without any prior training specific to the evaluation task.

\paragraph{(8)  Few-shot and One-shot Approaches:} These approaches involve leveraging a limited amount of training data to serve as exemplars to direct the model's responses. While we experimented with a one-shot setting, it is important to note that increasing the number of examples in the prompt had the unintended consequence of slowing down inference.

\begin{table*}[h]
\centering
\begin{tabular}{|p{15cm}|}
\hline
1. Coherence: Assess how well the summary conveys a clear and logical message.\\
2. Consistency: Check if the summary accurately represents the main points of the news article.\\
3. Fluency: Evaluate the smoothness and readability of the summary.\\
4. Relevance: Determine if the summary is relevant to the news article's topic.\\
For each sentence in the summary, assign a score from 1 to 5 for each aspect (coherence, consistency, fluency, and relevance).\\

Example:\\
Sentence 1: ``The company announced a new product line.''\\
Coherence: 4\\
Consistency: 3\\
Fluency: 3\\
Relevance: 4\\
\color{blue}{Total Score: {(Coherence + Consistency + Fluency + Relevance) / 4}}\\

Total Score: $ (4 + 3 + 3 + 4) / 4 = 14 / 4 = 3.5$ \\
So, the summary has an overall score of $3.5$ out of $5$.\\
\hline
\end{tabular}
\caption{
Example of a prompt generated for summarization quality estimation. These instructions demonstrate step-by-step, with the aid of an example, how the final score should be calculated.}
\label{tab-cot1}
\end{table*}

\begin{table}[H]
\centering
\begin{tabular}{|p{7.4 cm}|}
\hline
{\color{blue}Example:\newline
1. Read the news article: ``A new study found that regular exercise can significantly improve mental health.'' \newline
2. Read the summary: ``A study discovered that exercise has a significant impact on mental health.''}\newline
3. Evaluate the summary based on the aspects:\newline
a. Coherence: 5 (The summary maintains a clear and logical flow of ideas.) \newline
b. Consistency: 5 (The main points of the news article are accurately represented.) \newline
c. Fluency: 5 (The summary is written in a smooth and easy-to-understand manner.)\newline
d. Relevance: 5 (The summary conveys the essential information from the news article.)\newline
4. Assign scores for each aspect:
[$\hdots$]\\
\hline
\end{tabular}
\caption{Example of a chain-of-thought prompt generated for summarization quality estimation. In this chain of thought prompt, descriptions are generated for each of the four aspects (coherence, consistency, fluency, and relevance).}
\label{tab-cot2}
\end{table}

\section{Results}
Table \ref{table:summarizationResults} shows results for a battery of approaches.  One should note that we tried many combinations of the approaches with different seed prompts, and the number of experiments is quite large.  For the sake of simplicity, we report on key combinations that we uncovered.  Prompt ID refers to the specific prompt that was used and the exact text can be found in the Appendix.  

\begin{table*}[h]

\centering
\begin{tabular}{|p{0.7in}|p{4.4in}|l|}
\hline
\textbf{Prompt ID} & \textbf{Approach} & \textbf{Score} \\
\hline
\textbf{P1} & (1) Standard Prompt  (7) Zero-shot & \textbf{0.3211} \\\hline
P2 & (1) Standard Prompt (7) Zero-shot  & 0.3075 \\\hline
P12 & (2) Annotator Instructions (6) CoT (7) Zero-shot  & 0.2837 \\ \hline
P14 & (3) LLM-generated (6)CoT (7) Zero-shot & 0.2827  \\\hline
P10 & (3) LLM-generated (6)CoT (8) One-shot & 0.2687 \\ \hline
P6 & (1) Standard Prompt (7) Zero-shot  & 0.2597 \\ \hline
P9 & (6) CoT (7) Zero-shot & 0.2477 \\ \hline
P11 & (3) LLM-generated (6)CoT (8) One-shot& 0.2245 \\ \hline
P13 & (6) CoT (8) One-shot& 0.2244 \\ \hline
P5 & (1) Standard Prompt   (7) Zero-shot & -0.0163 \\ \hline
P4 &(1) Standard Prompt (7) Zero-shot  & -0.0172 \\ \hline
P7 & (1) Standard Prompt (7) Zero-shot & -0.0209 \\ \hline
P3 & (1) Standard Prompt (7) Zero-shot & -0.0255 \\ \hline
P8 &(1) Standard Prompt (7) Zero-shot  & -0.0329 \\ \hline
Baseline & Direct Assessment   & 0.3065 \\\hline
Baseline & Random   & -0.0340 \\
\hline
\end{tabular}
\caption{ Results of prompts for evaluating summarization. Score is the dev score obtained from the Codalab submission. The Prompt ID map to the full Prompts in the Appendix. The details for each of the Approaches can be found in Section 3.}
\label{table:summarizationResults}
\end{table*}

Among the approaches, Prompt P1, which employs a standard manual prompt in a zero-shot setting with a grading scale ranging from 1 to 5, emerges as the top performer, achieving a notable score of 0.3211 
on the development dataset.  This was surprising as this is essentially one of the most straightforward approaches to try.  However, it is perhaps unsurprising as the Direct Assessment baseline provided by the Shared Task is also a simple manual prompt in a zero-shot setting~\cite{kocmi-federmann-2023-large}.  

Following closely behind, we find approaches that leverage annotator instructions as seed prompts (P12) and prompts generated using LLMs (P14 and P10), all in zero-shot settings, also stand out. These prompts, in contrast to standard ones, contain a wealth of details about the evaluation metrics, offering intricate guidance to the model.

However, our exploration of a one-shot setting indicates that this approach does not yield as promising results (as much as 0.1 behind our best approach combination). Further experimentation with the choice of examples provided to the model may be warranted to enhance its performance.

Notably, the incorporation of chain-of-thought prompting appears to be a beneficial strategy, as evidenced by the strong performance observed (our third, fourth, and fifth-best experiments).

On the lower end of the result scores, we find standard prompts (P5, P4, P7, P3, P8) that utilize grading scales with exceptionally high precision or qualitative labels as quality measures. These prompts, while designed with attention to detail, exhibit comparatively lower scores in the evaluation, suggesting the importance of striking a balance between precision and other factors when designing prompts for quality estimation tasks. 

Prompt P6 achieves the highest Kendall correlation coefficient on the test set with a score of 0.4423. Furthermore, P1 and P2 also achieve competitive scores of 0.4419 and 0.4422 respectively.

\section{Discussion}

The evaluation results we have presented furnish compelling evidence regarding the proficiency of LLMs in the domain of quality estimation for summarization. Our findings underscore the capacity of these models to provide valuable insights into the quality of generated outputs, thereby contributing to the advancement of evaluation methodologies within the field of NLP.  

\subsection{Insights}


\textbf{Scoring Rubric Matters:} Our experimentation with scoring rubrics revealed an intriguing trend. While assessing the precision and range of scores requested from the model, we observed that quantitative scores with lower precision exhibited favorable performance. Surprisingly, the use of qualitative labels such as ``Very Poor'', ``Poor'', ``Average'', ``Good'', and ``Very Good'' to describe quality yielded comparatively less favorable results as can be seen from the dev scores in Table \ref{table:scoreRubrics}. This suggests that when instructing LLMs for quality estimation, a preference for quantitative, less granular scoring may be more effective.

\begin{table*}
\centering
\begin{tabular}{|p{5.5in}|p{0.3in}|}
\hline
\centering\textbf{Prompt*} & \textbf{Dev score}  
\\
\hline
Provide an overall score for the quality of this summary in the range {\color{blue} 1 (worst) to 5 (best).} Possible scores are {\color{magenta}1, 1.5, 2, 2.5, 3, 3.5, 4, 4.5 and 5.} & 0.32 
\\\hline
Provide an overall score for the quality of this summary in the range {\color{blue} 0 (worst) to 100 (best).} Possible scores are {\color{magenta}0, 10, 20, 30, 40, 50, 60, 70, 80, 90 and 100.} & 0.31 
\\\hline
Provide an overall score for the quality of this summary in the range {\color{blue}1 (worst) to 5 (best)} that is an {\color{magenta}average of the scores (also from 1 to 5)} for fluency, coherence, consistency and relevance.& -0.03 
\\\hline
Provide an overall score for the quality of this summary in the range {\color{blue}0 (worst) to 100 (best)}. Possible scores are {\color{magenta}0, 5, 10, 15, 20, 25, 30, 35, 40, 45, 50, 55, 60, 65, 70, 75, 80, 85, 90, 95 and 100.}& -0.02 
\\\hline
Provide an overall score for the quality of this summary in the range {\color{blue}Very Poor (worst) to Very Good (best).} Possible scores are {\color{magenta}Very Poor, Poor, Average, Good, Very Good.} & -0.02 
\\\hline
Provide an overall score for the quality of this summary in the range {\color{blue}Incomprehensible (worst) to Excellent (best).} Possible scores are {\color{magenta}Incomprehensible, Poor, Average, Good, Excellent.} & -0.03 
\\
\hline
\end{tabular}
\caption{
Standard prompting with different score ranges*The prompts in the table are appended to \textit{Given the following summary for a news article, evaluate this summary for its fluency, coherence, consistency and relevance.} along with the \textit{input} to form the full instruction to the model.
}

\label{table:scoreRubrics}
\end{table*}

\textbf{Effect on Performance through Examples:} To enhance performance, we incorporated explicit examples into zero-shot prompts for each score on the evaluation scale. Contrary to our expectations, the inclusion of examples did not yield a noticeable improvement in model performance. This observation highlights the nuanced nature of prompt design and underscores the need for tailored approaches that align with the unique characteristics of the task.

\textbf{Simpler Prompts Suffice:}
A notable finding emerged from our exploration of prompt complexity. 
While we originally hypothesized that detailed prompts derived from SummEval annotation guidelines would outperform simpler prompts based on the same four quality dimensions (fluency, coherence, consistency, relevance), our results did not substantiate this hypothesis. However, it is worth noting that this approach shows promise, particularly when the aspect being evaluated is ambiguous to the model. The provision of detailed prompts with examples and context holds the potential to improve performance, suggesting a fruitful direction for future research.



\subsection{Hallucinations during prompt refinement}
We conducted an experiment where we leveraged a separate LLM to generate instructions for assessing the quality of summarizations.
This approach yielded promising results initially, as the generated instructions were detailed and exhibited potential when tested with a subset of examples from the training dataset. However, during our extensive evaluation process, we encountered instances of hallucinations within the generated instructions, prompting us to incorporate further refinement steps into our methodology. Hallucinations in this context refer to the phenomenon where the model produces content in the instruction that deviates from the original intent or context, thereby introducing inaccuracies or inconsistencies \cite{maynez-etal-2020-faithfulness}.

One recurring form of hallucination pertained to the numbering within the instructions. In particular, we observed instances where the numbering in the latter part of the instruction was hallucinated, leading to inconsistencies. For instance, as shown in Table ~\ref{table:hallucinations}, the numbering sequence (2, 3, 4, 4) in the model's response was incongruent with the original (1, 2, 3, 4) prompt. Additionally, we identified another hallucination wherein the phrase ``Answer:'' was erroneously introduced in the model's output. This unintended addition in the response further underscores the need for vigilance when using LLMs to generate instructions.  For our experiments, we manually removed LLM-generate prompts with hallucinations.


 \begin{table*}
\centering
\begin{tabular}{p{3in}|p{3in}}

\centering\textbf{Response without hallucinations} & \textbf{Response with hallucinations} \\
\hline
1. Relevance: 2 Explanation: The summary does not capture the key points of the article. It only mentions the man gently cupping the fish, but it does not mention the relationship between the man and the fish, the video, or the fish's behavior. \newline 2. Consistency: 2 Explanation: The summary does not reproduce all facts accurately. It does not mention the fish's hybrid origin or the bullying of the other fish.\newline 3. Fluency: 2  Explanation: The summary is not well-written and grammatically correct. It contains grammatical errors and is not easy to understand. \newline 4. Coherence: 2 Explanation: The summary does not sound naturally and collectively. It is difficult to understand and does not flow well as a whole.&
{\color{red}2.} Relevance: 3 Explanation: The summary captures the key points of the article, but it is missing some important aspects, such as the half-and-half scarves and the fact that the match was played at Altrincham's Moss Lane. {\color{red}Answer:} \newline {\color{red}3.} Consistency: 3 Explanation: The summary does reproduce all facts accurately, but it makes up untrue information by stating that the match was played at Old Trafford. {\color{red}Answer:} \newline {\color{red}4.} Fluency: 3 Explanation: The individual sentences are well-written and grammatically correct, but the summary could be more coherent and sound more naturally. {\color{red}Answer:} \newline {\color{red}4.} Coherence: 3 Explanation: The summary is not as coherent as it could be, as it jumps from one topic to another without a clear flow. \\
\end{tabular}
\caption{ Hallucinations observed in prompts that are refined using LLMs.
}
\label{table:hallucinations}
\end{table*}

\section{Conclusion}
In this paper, we have presented our system entry for the Eval4NLP 2023 Shared Task on Prompting Large Language Models as Explainable Metrics. Our approach revolves around assessing the capabilities of the small \verb|orca_mini_v3_7B| model in the reference-free evaluation of generated summaries. Through a series of extensive experiments, we have explored the model's adaptability and responsiveness to various prompting techniques, including zero-shot, one-shot, and chain-of-thought prompts. These experiments were conducted using meticulously designed prompts that emphasize key evaluation criteria.
Our findings suggest that the use of a standard prompting (when using a small model) is performant while the chain-of-thought prompt-based approach holds significant promise for evaluating the quality of generated summaries.  However, we do note that slight variations in the underlying prompt text could swing performances rather widely, so more research is definitely required to understand this phenomenon.

\bibliography{anthology, custom}

\begin{thebibliography}{17}
\expandafter\ifx\csname natexlab\endcsname\relax\def\natexlab#1{#1}\fi

\bibitem[{Brown et~al.(2020)Brown, Mann, Ryder, Subbiah, Kaplan, Dhariwal, Neelakantan, Shyam, Sastry, Askell, Agarwal, Herbert-Voss, Krueger, Henighan, Child, Ramesh, Ziegler, Wu, Winter, Hesse, Chen, Sigler, Litwin, Gray, Chess, Clark, Berner, McCandlish, Radford, Sutskever, and Amodei}]{llmfewshot}
Tom Brown, Benjamin Mann, Nick Ryder, Melanie Subbiah, Jared~D Kaplan, Prafulla Dhariwal, Arvind Neelakantan, Pranav Shyam, Girish Sastry, Amanda Askell, Sandhini Agarwal, Ariel Herbert-Voss, Gretchen Krueger, Tom Henighan, Rewon Child, Aditya Ramesh, Daniel Ziegler, Jeffrey Wu, Clemens Winter, Chris Hesse, Mark Chen, Eric Sigler, Mateusz Litwin, Scott Gray, Benjamin Chess, Jack Clark, Christopher Berner, Sam McCandlish, Alec Radford, Ilya Sutskever, and Dario Amodei. 2020.
\newblock \href {https://proceedings.neurips.cc/paper_files/paper/2020/file/1457c0d6bfcb4967418bfb8ac142f64a-Paper.pdf} {Language models are few-shot learners}.
\newblock In \emph{Advances in Neural Information Processing Systems}, volume~33, pages 1877--1901. Curran Associates, Inc.

\bibitem[{Camburu et~al.(2018)Camburu, Rockt{\"a}schel, Lukasiewicz, and Blunsom}]{camburu2018snli}
Oana-Maria Camburu, Tim Rockt{\"a}schel, Thomas Lukasiewicz, and Phil Blunsom. 2018.
\newblock e-snli: Natural language inference with natural language explanations.
\newblock \emph{Advances in Neural Information Processing Systems}, 31.

\bibitem[{Chiang and Lee(2023)}]{chiang-lee-2023-large}
Cheng-Han Chiang and Hung-yi Lee. 2023.
\newblock \href {https://doi.org/10.18653/v1/2023.acl-long.870} {Can large language models be an alternative to human evaluations?}
\newblock In \emph{Proceedings of the 61st Annual Meeting of the Association for Computational Linguistics (Volume 1: Long Papers)}, pages 15607--15631, Toronto, Canada. Association for Computational Linguistics.

\bibitem[{Fabbri et~al.(2020)Fabbri, Kry{\'s}ci{\'n}ski, McCann, Xiong, Socher, and Radev}]{fabbri2020summeval}
Alexander~R Fabbri, Wojciech Kry{\'s}ci{\'n}ski, Bryan McCann, Caiming Xiong, Richard Socher, and Dragomir Radev. 2020.
\newblock Summeval: Re-evaluating summarization evaluation.
\newblock \emph{arXiv preprint arXiv:2007.12626}.

\bibitem[{Freitag et~al.(2022)Freitag, Rei, Mathur, Lo, Stewart, Avramidis, Kocmi, Foster, Lavie, and Martins}]{freitag-etal-2022-results}
Markus Freitag, Ricardo Rei, Nitika Mathur, Chi-kiu Lo, Craig Stewart, Eleftherios Avramidis, Tom Kocmi, George Foster, Alon Lavie, and Andr{\'e} F.~T. Martins. 2022.
\newblock \href {https://aclanthology.org/2022.wmt-1.2} {Results of {WMT}22 metrics shared task: Stop using {BLEU} {--} neural metrics are better and more robust}.
\newblock In \emph{Proceedings of the Seventh Conference on Machine Translation (WMT)}, pages 46--68, Abu Dhabi, United Arab Emirates (Hybrid). Association for Computational Linguistics.

\bibitem[{Fu et~al.(2023)Fu, Ng, Jiang, and Liu}]{DBLP:journals/corr/abs-2302-04166}
Jinlan Fu, See{-}Kiong Ng, Zhengbao Jiang, and Pengfei Liu. 2023.
\newblock \href {https://doi.org/10.48550/arXiv.2302.04166} {Gptscore: Evaluate as you desire}.
\newblock \emph{CoRR}, abs/2302.04166.

\bibitem[{Kocmi and Federmann(2023)}]{kocmi-federmann-2023-large}
Tom Kocmi and Christian Federmann. 2023.
\newblock \href {https://aclanthology.org/2023.eamt-1.19} {Large language models are state-of-the-art evaluators of translation quality}.
\newblock In \emph{Proceedings of the 24th Annual Conference of the European Association for Machine Translation}, pages 193--203, Tampere, Finland. European Association for Machine Translation.

\bibitem[{Kotonya and Toni(2020)}]{kotonya-toni-2020-explainable}
Neema Kotonya and Francesca Toni. 2020.
\newblock \href {https://doi.org/10.18653/v1/2020.coling-main.474} {Explainable automated fact-checking: A survey}.
\newblock In \emph{Proceedings of the 28th International Conference on Computational Linguistics}, pages 5430--5443, Barcelona, Spain (Online). International Committee on Computational Linguistics.

\bibitem[{Kryscinski et~al.(2019)Kryscinski, Keskar, McCann, Xiong, and Socher}]{kryscinski-etal-2019-neural}
Wojciech Kryscinski, Nitish~Shirish Keskar, Bryan McCann, Caiming Xiong, and Richard Socher. 2019.
\newblock \href {https://doi.org/10.18653/v1/D19-1051} {Neural text summarization: A critical evaluation}.
\newblock In \emph{Proceedings of the 2019 Conference on Empirical Methods in Natural Language Processing and the 9th International Joint Conference on Natural Language Processing (EMNLP-IJCNLP)}, pages 540--551, Hong Kong, China. Association for Computational Linguistics.

\bibitem[{Leiter et~al.(2023)Leiter, Opitz, Deutsch, Gao, Dror, and Eger}]{eval4nlp23}
Christoph Leiter, Juri Opitz, Daniel Deutsch, Yang Gao, Rotem Dror, and Steffen Eger. 2023.
\newblock The eval4nlp 2023 shared task on prompting large language models as explainable metrics.
\newblock In \emph{Proceedings of the 4th Workshop on Evaluation and Comparison for NLP systems}.

\bibitem[{Liu et~al.(2023)Liu, Iter, Xu, Wang, Xu, and Zhu}]{DBLP:journals/corr/abs-2303-16634}
Yang Liu, Dan Iter, Yichong Xu, Shuohang Wang, Ruochen Xu, and Chenguang Zhu. 2023.
\newblock \href {https://doi.org/10.48550/arXiv.2303.16634} {G-eval: {NLG} evaluation using {GPT-4} with better human alignment}.
\newblock \emph{CoRR}, abs/2303.16634.

\bibitem[{Maynez et~al.(2020)Maynez, Narayan, Bohnet, and McDonald}]{maynez-etal-2020-faithfulness}
Joshua Maynez, Shashi Narayan, Bernd Bohnet, and Ryan McDonald. 2020.
\newblock \href {https://doi.org/10.18653/v1/2020.acl-main.173} {On faithfulness and factuality in abstractive summarization}.
\newblock In \emph{Proceedings of the 58th Annual Meeting of the Association for Computational Linguistics}, pages 1906--1919, Online. Association for Computational Linguistics.

\bibitem[{Mendon{\c{c}}a et~al.(2023)Mendon{\c{c}}a, Pereira, Moniz, Paulo~Carvalho, Lavie, and M~Trancoso}]{mendonca-etal-2023-simple}
John Mendon{\c{c}}a, Patr{\'\i}cia Pereira, Helena Moniz, Joao Paulo~Carvalho, Alon Lavie, and Isabel M~Trancoso. 2023.
\newblock \href {https://aclanthology.org/2023.dstc-1.16} {Simple {LLM} prompting is state-of-the-art for robust and multilingual dialogue evaluation}.
\newblock In \emph{Proceedings of The Eleventh Dialog System Technology Challenge}, pages 133--143, Prague, Czech Republic. Association for Computational Linguistics.

\bibitem[{Wei et~al.(2022)Wei, Wang, Schuurmans, Bosma, ichter, Xia, Chi, Le, and Zhou}]{chainofthought}
Jason Wei, Xuezhi Wang, Dale Schuurmans, Maarten Bosma, brian ichter, Fei Xia, Ed~Chi, Quoc~V Le, and Denny Zhou. 2022.
\newblock \href {https://proceedings.neurips.cc/paper_files/paper/2022/file/9d5609613524ecf4f15af0f7b31abca4-Paper-Conference.pdf} {Chain-of-thought prompting elicits reasoning in large language models}.
\newblock In \emph{Advances in Neural Information Processing Systems}, volume~35, pages 24824--24837. Curran Associates, Inc.

\bibitem[{Xu et~al.(2023)Xu, Wang, Pan, Song, Freitag, Wang, and Li}]{xu2023instructscore}
Wenda Xu, Danqing Wang, Liangming Pan, Zhenqiao Song, Markus Freitag, William~Yang Wang, and Lei Li. 2023.
\newblock \href {http://arxiv.org/abs/2305.14282} {Instructscore: Towards explainable text generation evaluation with automatic feedback}.

\bibitem[{Yang et~al.(2023)Yang, Zhang, Tao, Wang, Wei, and Jiang}]{yang2023knowledgeprompted}
Hao Yang, Min Zhang, Shimin Tao, Minghan Wang, Daimeng Wei, and Yanfei Jiang. 2023.
\newblock \href {http://arxiv.org/abs/2306.07486} {Knowledge-prompted estimator: A novel approach to explainable machine translation assessment}.

\bibitem[{Zellers et~al.(2019)Zellers, Holtzman, Rashkin, Bisk, Farhadi, Roesner, and Choi}]{zellers2019defending}
Rowan Zellers, Ari Holtzman, Hannah Rashkin, Yonatan Bisk, Ali Farhadi, Franziska Roesner, and Yejin Choi. 2019.
\newblock Defending against neural fake news.
\newblock \emph{Advances in neural information processing systems}, 32.

\end{thebibliography}

\appendix

\section{Software Libraries and Hardware}

We use PyTorch v2.0.1 compiled with CUDA 11.7. We use version 0.0.64 of the guidance library\footnote{\url{https://pypi.org/project/guidance/}} for prompt generation and prompt evaluation. We conduct all experiments on NVIDIA T4 and NVIDIA A10G GPUs.


\clearpage

\onecolumn
\section{Prompt Templates} 

\subsection{Summarization}
Below we provide our prompt templates which we use for the experiments described in this paper.

\begin{longtable}{|p{0.5in}|p{5.5in}|}
\hline
\centering\textbf{Prompt ID} & \textbf{Prompt} \\
\hline
P1& Given the following summary for a news article, evaluate this summary for its fluency, coherence, consistency and relevance. Provide an overall score for the quality of this summary in the range 1 (worst) to 5 (best). Possible scores are 1, 1.5, 2, 2.5, 3, 3.5, 4, 4.5 and 5.\\\hline
P2&Given the following summary for a news article, evaluate this summary for its fluency, coherence, consistency and relevance. Provide an overall score for the quality of this summary in the range 0 (worst) to 100 (best). Possible scores are 0, 10, 20, 30, 40, 50, 60, 70, 80, 90 and 100.\\\hline
P3&Given the following summary for a news article, evaluate this summary for its fluency, coherence, consistency and relevance. Provide an overall score for the quality of this summary in the range 1 (worst) to 5 (best) that is an average of the scores (also from 1 to 5) for fluency, coherence, consistency and relevance.\\\hline
P4&Given the following summary for a news article, evaluate this summary for its fluency, coherence, consistency and relevance. Provide an overall score for the quality of this summary in the range 0 (worst) to 100 (best). Possible scores are 0, 5, 10, 15, 20, 25, 30, 35, 40, 45, 50, 55, 60, 65, 70, 75, 80, 85, 90, 95 and 100.\\\hline
P5&Given the following summary for a news article, evaluate this summary for its fluency, coherence, consistency and relevance. Provide an overall score for the quality of this summary in the range 0 (worst) to 100 (best). Possible scores are 0, 1, 2, 3, 4, 5, 6, 7, 8, 9, 10, 11, 12, 13, 14, 15, 16, 17, 18, 19, 20, 21, 22, 23, 24, 25, 26, 27, 28, 29, 30, 31, 32, 33, 34, 35, 36, 37, 38, 39, 40, 41, 42, 43, 44, 45, 46, 47, 48, 49, 50, 51, 52, 53, 54, 55, 56, 57, 58, 59, 60, 61, 62, 63, 64, 65, 66, 67, 68, 69, 70, 71, 72, 73, 74, 75, 76, 77, 78, 79, 80, 81, 82, 83, 84, 85, 86, 87, 88, 89, 90, 91, 92, 93, 94, 95, 96, 97, 98, 99, 100.\\\hline
P6&Given the following summary for a news article, evaluate this summary for its fluency, coherence, consistency and relevance. Provide an overall score for the quality of this summary in the range -100 (worst) to 100 (best). Possible scores are -100, -50, 0, 50, 100.\\\hline
P7&Given the following summary for a news article, evaluate this summary for its fluency, coherence, consistency and relevance. Provide an overall score for the quality of this summary in the range Very Poor (worst) to Very Good (best). Possible scores are Very Poor, Poor, Average, Good, Very Good.\\\hline
P8&Given the following summary for a news article, evaluate this summary for its fluency, coherence, consistency and relevance. Provide an overall score for the quality of this summary in the range Incomprehensible (worst) to Excellent (best). Possible scores are Incomprehensible, Poor, Average, Good, Excellent.\\\hline
P9& \verb|### System:| You are an AI assistant that follows instruction extremely well. Help as much as you can. \newline
\newline
 \verb|### User:|\newline
In this task you will evaluate the quality of summaries written for a news article\newline
To correctly solve this task, follow these steps:\newline
1. Carefully read the news articles, be aware of the information it contains.\newline
2. Read the proposed summary.\newline
3. Rate each summary on a scale from 1 (Worst) to 5 (Best) by its relevance, consistency, fluency, and coherence.\newline
Relevance: ``The rating measures how well the summary captures the key points of the article. Consider whether all and only the important aspects are contained in the summary.''\newline
Consistency: ``The rating measures the facts in the summary are consistent with the facts in the original article. Consider whether the summary does reproduce all facts accurately and does not make up untrue information.''\newline
Fluency: ``This rating measures the quality of individual sentences, are they well-written and grammatically correct. Consider the quality of individual sentences.''\newline
Coherence: ``The rating measures the quality of all sentences collectively, to the fit together and sound naturally. Consider the quality of the summary as a whole.''\newline
Format the response as follows: \newline
Answer: \newline
Relevance: \verb|<Rating for Relevance>|\newline
Explanation: \verb|<Evidence for Relevance rating>|\newline
Consistency: \verb|<Rating for Consistency>|\newline
Explanation: \verb|<Evidence for Consistency rating>|\newline
Fluency: \verb|<Rating for Fluency>|\newline
Explanation: \verb|<Evidence for Fluency rating>|\newline
Coherence: \verb|<Rating for Coherence>|\newline
Explanation: \verb|<Evidence for Coherence rating>|\newline
News article: \verb|{source_text}|\newline
Summary: \verb|{summary}|\newline
\newline
\verb|### Assistant:|\newline
Answer:\\\hline
P10& \verb|### User:| \newline
Evaluate the summary for a news article by assigning a score from 0 to 100 for each of the following aspects: Coherence, Consistency, Fluency, and Relevance. \newline
1. Coherence: \newline
- Read the summary and determine if it is well-structured, easy to understand, and logically connected. \newline
- Assign a score from 0 to 100 based on how well the summary is coherent. \newline
2. Consistency: \newline
- Check if the summary accurately represents the main points and ideas from the original news article.\newline
 - Assign a score from 0 to 100 based on how consistent the summary is with the original content.\newline
3. Fluency:\newline
- Evaluate the clarity and smoothness of the summary.\newline
- Assign a score from 0 to 100 based on how well the summary is written and easy to read.\newline
4. Relevance:\newline
- Determine if the summary effectively conveys the most important information from the original news article.\newline
- Assign a score from 0 to 100 based on how relevant and informative the summary is.\newline
Once you have evaluated each aspect, add up the scores and assign a final score from 0 to 100 for the overall summary.\newline
News article: \verb|{source_text}|\newline
Summary: \verb|{summary}|\newline\newline
\verb|### Assistant:|\newline
Evaluation:\\\hline
P11& \verb|### User|: \newline
1. Read the news article carefully. \newline
2. Read the summary of the news article.\newline
3. Evaluate the summary based on the following aspects:\newline
a. Coherence: How well does the summary maintain a clear and logical flow of ideas?\newline
b. Consistency: Does the summary accurately represent the main points of the news article?\newline
c. Fluency: Is the summary written in a smooth and easy-to-understand manner? \newline
d. Relevance: Does the summary convey the essential information from the news article? \newline
4. Assign a score from 1 to 5 for each aspect based on the evaluation. \newline
5. Average the scores for each aspect to get the overall score for the summary.\newline
Example:\newline
1. Read the news article: ``A new study found that regular exercise can significantly improve mental health.''\newline
2. Read the summary: ``A study discovered that exercise has a significant impact on mental health.''\newline
3. Evaluate the summary based on the aspects:\newline
a. Coherence: 5 (The summary maintains a clear and logical flow of ideas.)\newline
b. Consistency: 5 (The main points of the news article are accurately represented.)\newline
c. Fluency: 5 (The summary is written in a smooth and easy-to-understand manner.)\newline
d. Relevance: 5 (The summary conveys the essential information from the news article.)\newline
4. Assign scores for each aspect:\newline
a. Coherence: 5\newline
b. Consistency: 5\newline
c. Fluency: 5\newline
d. Relevance: 5\newline
5. Average the scores for each aspect:\newline
a. Coherence: 5\newline
b. Consistency: 5\newline
c. Fluency: 5\newline
d. Relevance: 5\newline
6. Average the scores for each aspect: 5\newline
7. Overall score for the summary: 5\newline
News article: \verb|{source_text}| \newline
Summary: \verb|{summary}| \newline
\verb|### Assistant:|\newline
Assign scores for each aspect: \\\hline
P12&\verb|### User|:\newline
Evaluate the summary for a news article by assigning a score from 0 to 100 for each of the following aspects: Coherence, Consistency, Fluency, and Relevance.\newline
1. Coherence:
- Read the summary and determine if it is well-structured, easy to understand, and logically connected.
- Assign a score from 0 to 100 based on how well the summary is coherent.\newline
2. Consistency:
- Check if the summary accurately represents the main points and ideas from the original news article.\newline
 - Assign a score from 0 to 100 based on how consistent the summary is with the original content.\newline
3. Fluency:
- Evaluate the clarity and smoothness of the summary.
- Assign a score from 0 to 100 based on how well the summary is written and easy to read.\newline
4. Relevance:
- Determine if the summary effectively conveys the most important information from the original news article.
- Assign a score from 0 to 100 based on how relevant and informative the summary is.\newline
Once you have evaluated each aspect, add up the scores and assign a final score from 0 to 100 for the overall summary.\newline
News article: \verb|{source_text}| \newline
Summary: \verb|{summary}|\newline
\verb|### Assistant|:\newline
Evaluation:\\\hline

P13& \verb|### User|:\newline
1. Coherence: Assess how well the summary conveys a clear and logical message. \newline
2. Consistency: Check if the summary accurately represents the main points of the news article. \newline
3. Fluency: Evaluate the smoothness and readability of the summary. \newline
4. Relevance: Determine if the summary is relevant to the news article's topic. \newline

For each sentence in the summary, assign a score from 1 to 5 for each aspect (coherence, consistency, fluency, and relevance).\newline

Example:\newline
Sentence 1: ``The company announced a new product line.''\newline
Coherence: 4\newline
Consistency: 3\newline
Fluency: 3\newline
Relevance: 4\newline

Total Score: (Coherence + Consistency + Fluency + Relevance) / 4\newline
Total Score: \texttt{(4 + 3 + 3 + 4) / 4 = 14 / 4 = 3.5} \newline

So, the summary has an overall score of 3.5 out of 5. \\\hline
P14& \verb|### User|:\newline
To evaluate the coherence, consistency, fluency and relevance of a summary of a news article, follow these steps:\newline
1. Read the news article carefully to understand its main points.\newline
2. Read the summary and place it next to the news article so that you can compare them.\newline
3. First, starts by evaluating the relevance of each sentence in the summary. Ask yourself if the sentence supports the main points of the news article and if it adds new and relevant information. You can assign a score to each sentence that evaluates this criterion and check whether it is coherent with the main points of the news article.\newline
4. Next, evaluate the consistency of each sentence in the summary with respect to the main points of the news article. Ask yourself if the sentence is consistent with the overall message of the news article. You can assign a score to each criterion and check whether it is consistent with the overall purpose of the news article.\newline
5. Start evaluating the fluency of each sentence. Check if each sentence flows well and if there are any awkward phrasing or sentence structures. You can assign a score to each sentence based on the fluency of the writing.\newline
6. Assign a score of between 0 and 100 to each aspect (coherence, consistency, fluency and relevance) to evaluate how well the summary accurately and clearly conveys the article’s content.\newline
For example, if the summary accurately conveys the article’s content and clearly and coherently communicates each aspect, you can assign a score of 100 for each aspect, but if it slightly or not at all conveys, you can assign a score of 0. \\\hline
\end{longtable}

\clearpage
\section{Explanations}
\label{appendix:explanations}

In Tables \ref{tab:prompt-explanation} and \ref{tab:prompt-explanation2}  present explanations generated for our experiments on the dev set. The prompt employed to generate the results in Tables \ref{tab:prompt-explanation} and \ref{tab:prompt-explanation2} is shown in Table \ref{tab:explainable-prompt}.

\begin{table*}[ht]
    \centering
    \begin{tabular}{|p{15cm}|}
        \hline
 \verb|### Answer:|  \\
2 (Worst)\\
Relevance: 2\\ 
Explanation: 
\textcolor{brickred}{{The summary contains some relevant information, but it is not comprehensive enough. It only mentions the guard slipping and falling, but it does not mention the incident being caught on camera or the reason for the slip.}}\\
Consistency: 2\\ Explanation: \textcolor{brickred}{{
The summary is consistent with the original article, but it does not reproduce all facts accurately. It does not mention the metal protectors nailed to the soles of the guard's boots, which is an important detail.}}\\
Fluency: 2\\  Explanation: \textcolor{brickred}{{
The summary is grammatically correct, but it is not well-written. It is difficult to understand and lacks proper sentence structure.}}\\
Coherence: 2 \\ Explanation:\textcolor{brickred}{{ The summary is coherent, but it is not well-structured. It is difficult to follow and does not flow naturally.}}\\
  
   \hline
    \end{tabular}

    \caption{Explanation and scores produced for a low-scoring summary. The explanations in the generated response are shown in \textcolor{brickred}{{red}}. Note that an overall assessment, \textit{2 (Worst)}, is also generated for this example.}
    
    \label{tab:prompt-explanation}
\end{table*}

\begin{table*}[ht]
    \centering
    \begin{tabular}{|p{15cm}|}
        \hline
 \verb|### Answer:|  \\
Relevance: 4\\
\textcolor{brickred}{Explanation: The summary captures the key points of the article, but it could be more detailed and specific.}\\
Consistency: 4\\
\textcolor{brickred}{Explanation: The summary accurately reproduces the main facts from the article, but it could be more detailed.}\\
Fluency: 3\\
\textcolor{brickred}{Explanation: The summary is grammatically correct, but it could be more concise and fluent.}\\
Coherence: 3\\
\textcolor{brickred}{Explanation: The summary is coherent, but it could be more natural and flowing.}\\
  
   \hline
    \end{tabular}

    \caption{Explanation and scores produced for a high-scoring summary from the dev set. The explanations in the generated response are shown in \textcolor{brickred}{{red}}.}
    
    \label{tab:prompt-explanation2}
\end{table*}

\begin{table}[h]
    \centering
    \begin{tabular}{|p{15cm}|}
    \hline
     \verb|### User:|  \\
In this task you will evaluate the quality of summaries written for a news article\\
To correctly solve this task, follow these steps:\\
1. Carefully read the news article, be aware of the information it contains.\\
2. Read the proposed summary.\\
3. Rate each summary on a scale from 1 (Worst) to 5 (Best) inclusive by its relevance, consistency, fluency, and coherence.\\
Relevance: "The rating measures how well the summary captures the key points of the article.
Consider whether all and only the important aspects are contained in the summary."\\
Consistency: "The rating measures the facts in the summary are consistent with the facts in the original article.
Consider whether the summary does reproduce all facts accurately and does not make up untrue information."\\
Fluency: "This rating measures the quality of individual sentences, are they well-written and grammatically correct.
Consider the quality of individual sentences."\\
Coherence: "The rating measures the quality of all sentences collectively, to the fit together and sound naturally. Consider the quality of the summary as a whole."\\
Format the response as follows: \\
Answer: \\
Relevance: $\langle$Rating for Relevance$\rangle$\\
Explanation:  $\langle$Evidence for Relevance rating$\rangle$\\
Consistency:  $\langle$Rating for Consistency$\rangle$\\
Explanation:  $\langle$Evidence for Consistency rating$\rangle$\\
Fluency:  $\langle$Rating for Fluency$\rangle$\\
Explanation:  $\langle$Evidence for Fluency rating$\rangle$\\
Coherence:  $\langle$Rating for Coherence$\rangle$\\
Explanation:  $\langle$Evidence for Coherence rating$\rangle$\\
News article: \texttt{\{source\_text\}} \\
Summary: \texttt{\{summary\}}   \\
     \verb|### Assistant:|  \\
    \hline
    \end{tabular}
    \caption{Instructions that prompt the LLM to generate explanations in addition to quality scores for a summary.}
    \label{tab:explainable-prompt}
\end{table}

\end{document}